# Advance and Refinement: The Evolution of UAV Detection and Classification Technologies

--Manuscript Draft--

| | |
|---|---|
| **Manuscript Number:** | IJCIP-D-24-00101 |
| **Article Type:** | Review article |
| **Keywords:** | UAV Detection; UAV Classification; Radar Technology; Sensor Fusion; Optical Sensor; Acoustic Sensor. |
| **Corresponding Author:** | Ildar Kurmashev, Candidate of Technical Sciences<br>Manash Kozybayev North Kazakhstan University<br>Petropavlovsk, North Kazakhstan region KAZAKHSTAN |
| **First Author:** | Vladislav Semenyuk |
| **Order of Authors:** | Vladislav Semenyuk |
| | Ildar Kurmashev, Candidate of Technical Sciences |
| | Alberto Lupidi, Prof. |
| | Dmitriy Alyoshin |
| | Liliya Kurmasheva |
| | Alessandro Cantelli-Forti, Prof. |
| **Abstract:** | This review provides a detailed analysis of the advancements in unmanned aerial vehicle (UAV) detection and classification systems from 2020 to today. It covers various detection methodologies such as radar, radio frequency, optical, and acoustic sensors, and emphasizes their integration via sophisticated sensor fusion techniques. The fundamental technologies driving UAV detection and classification are thoroughly examined, with a focus on their accuracy and range. Additionally, the paper discusses the latest innovations in artificial intelligence and machine learning, illustrating their impact on improving the accuracy and efficiency of these systems. The review concludes by predicting further technological developments in UAV detection, which are expected to enhance both performance and reliability. |
| **Suggested Reviewers:** | Walter Matta<br>University of Rome<br>w.matta@unilink.it |
| | Amerigo Capria<br>Consorzio Nazionale Interuniversitario per le Telecomunicazioni<br>amerigo.capria@cnit.it |
| | Fabrizio Berizzi<br>University of Pisa<br>w.matta@unilink.it |
| **Opposed Reviewers:** | |

# Advance and Refinement: The Evolution of UAV Detection and Classification Technologies

Vladislav Semenyuk[a], Ildar Kurmashev[a], Alberto Lupidi[b], Dmitriy Alyoshin[a], Liliya Kurmasheva[a], Alessandro Cantelli-Forti[b]

[a]*M. Kozybaev North Kazakhstan University, Pushkin street, 86, Petropavlovsk, 150000, Kazakhstan*

[b]*RaSS (Radar and Surveillance Systems) National Laboratory, 56124 - Pisa, Italy*

**Abstract**

This review provides a detailed analysis of the advancements in unmanned aerial vehicle (UAV) detection and classification systems from 2020 to today. It covers various detection methodologies such as radar, radio frequency, optical, and acoustic sensors, and emphasizes their integration via sophisticated sensor fusion techniques. The fundamental technologies driving UAV detection and classification are thoroughly examined, with a focus on their accuracy and range. Additionally, the paper discusses the latest innovations in artificial intelligence and machine learning, illustrating their impact on improving the accuracy and efficiency of these systems. The review concludes by predicting further technological developments in UAV detection, which are expected to enhance both performance and reliability.

## 1. Introduction

Unmanned aerial vehicles (UAVs), widely used in various fields including military operations and commercial delivery, pose a significant security threat due to potential misuse for espionage or terrorist attacks. To counter these threats, effective long-distance UAV detection technologies are crucial. Key methods include radio frequency (RF)-based systems, radar, and both acoustic and optoelectronic detection, ensuring reliable identification and security measures against UAV-related risks.

Radar recognition is a prevalent method for detecting and tracking UAVs (Unmanned Aerial Vehicles), leveraging radar signals to pinpoint the UAV’s location and velocity. The efficacy of radar systems in UAV detection hinges on several factors: the UAV’s size, its flying altitude, the specific characteristics of the radar, as well as environmental elements like weather conditions, terrain, and surrounding reflective objects. However, radar systems face challenges such as clutter and radio interference, which can diminish detection range and accuracy in target location. The positioning of radar installations at higher elevations helps extend their detection range by reducing obstructions from ground-level objects [1-3].

RF (Radio Frequency) recognition analyzes electromagnetic signals emitted by the UAV, offering the capability to track UAVs over extended distances. Despite its advantages, this method requires sophisticated equipment and is prone to interference. One of its main drawbacks is its limited ability to pinpoint the exact direction of the UAV without extensive observation, which improves accuracy over time. Furthermore, RF systems often inaccurately estimate the range and altitude of the target [1-3].

Optoelectronic recognition systems utilize cameras and optical sensors to detect and monitor UAVs. This approach is notably precise and versatile across various settings, though it depends heavily on adequate lighting conditions. Optoelectronic systems' performance is influenced by the time of day and weather conditions like haze or precipitation, which can impair visibility and reduce detection efficiency. In contrast to radar systems, optoelectronic methods generally have a shorter UAV detection range but provide superior target recognition capabilities [4-5].

Acoustic recognition systems, which analyze sound emissions from UAV engines, are effective for short-range detection. These systems, however, are less reliable in urban or noisy settings and typically offer lower accuracy in locating UAVs, especially smaller models.

Recent advancements in data analysis and machine learning have led to the development of Sensor Fusion systems, which integrate these diverse detection technologies to enhance the accuracy and classification of UAVs as targets. Figure 1 illustrates the Sensor Fusion system architecture. The subsequent sections will delve into the technologies introduced by researchers for UAV detection and recognition from 2020 to the present, highlighting significant contributions and innovations in the field [1-5].

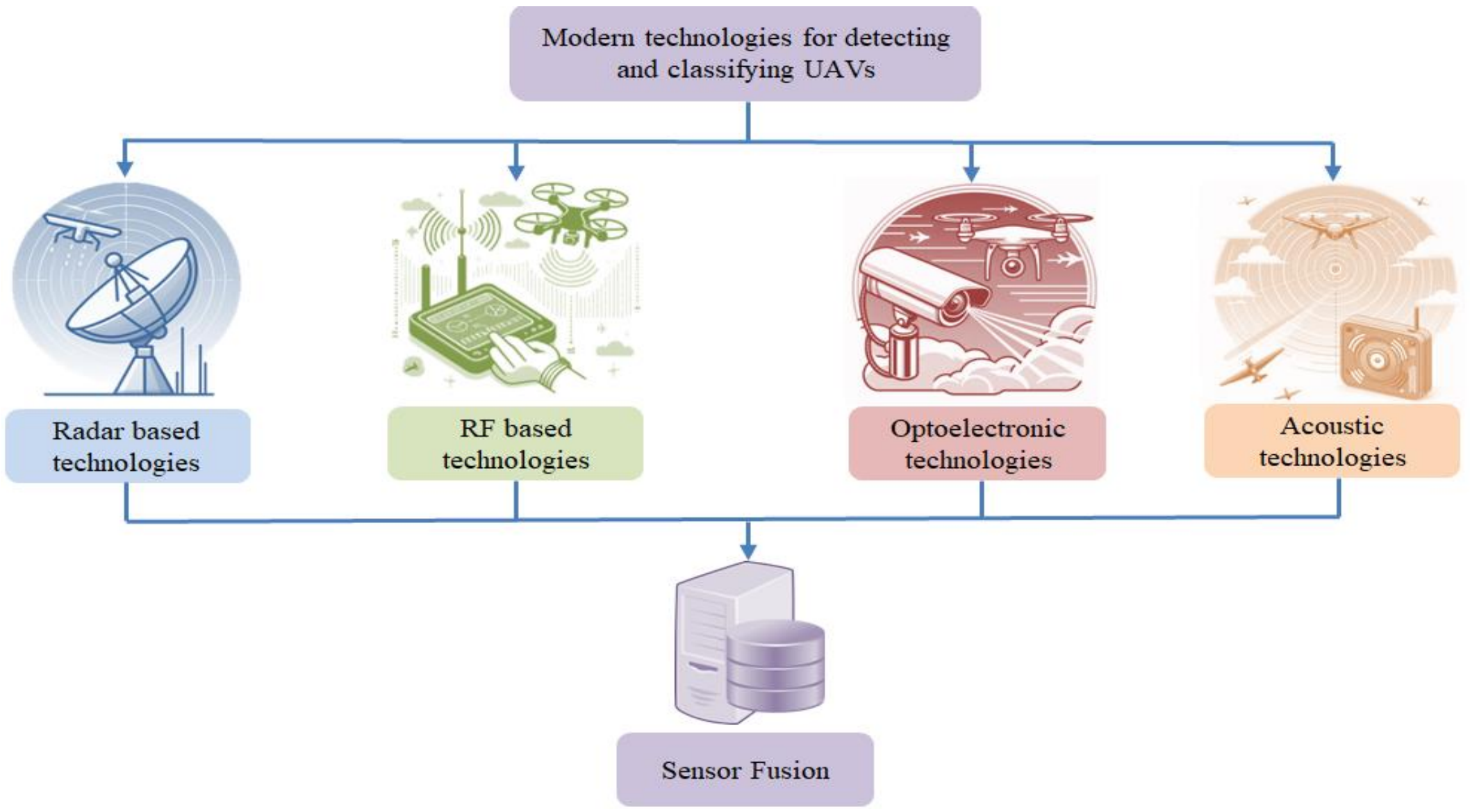

Fig. 1. Diagram of modern UAV detection and classification systems.

## 2. UAV detection and recognition technologies

*2.1. The toolbar and its menus Technologies for detecting and recognizing UAVs based on radar systems*

All technologies characteristic of UAV recognition radar systems can be divided into two classifications: by frequency range and by type of modulation (Fig. 2).

Classification by frequency range includes:

- L band [8, 10]
- S band [7, 10, 11]
- C band [12]
- X band [10]
- Ka band [6]
- W band [12]

Classification by type of modulation:

- FMCW (Frequency-Modulated Continuous-Wave) [9, 11, 12, 15, 19]
- LFPM (Linear Frequency-Phase Modulation) [10]
- FSK (Frequency Shift Keying) [7]
- LFM (Linear Frequency Modulation) [7]
- SFM (Stepped Frequency Modulation) [7]

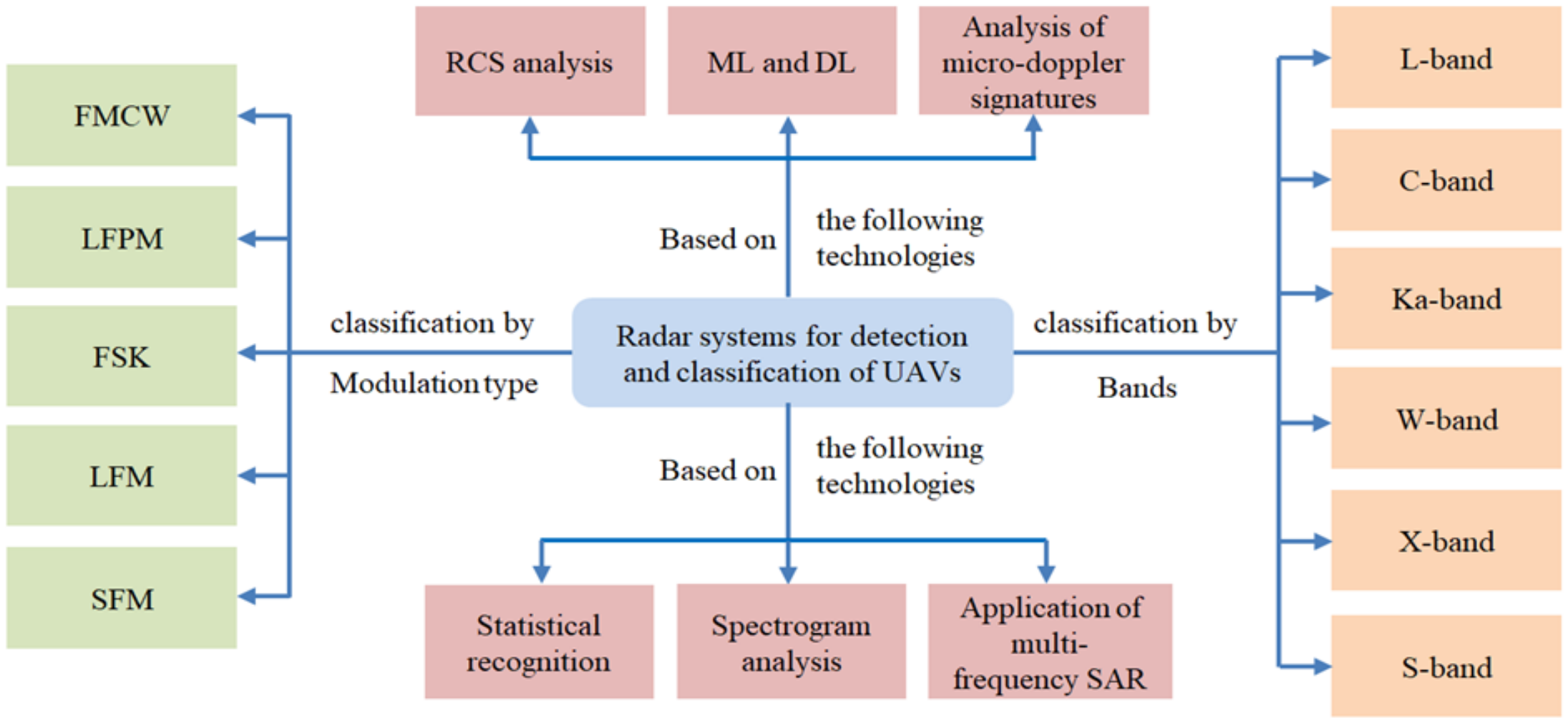

Fig. 2. Diagram of radar systems for detecting and classifying UAVs.

In UAV radar recognition systems, a number of technologies can be distinguished that are combined with each other for the purpose of detection and classification:

1. Radar Cross Section (RCS) analysis: Measurement and analysis of the radar cross section of UAVs for their classification and identification [6, 9].
2. Machine learning and Deep learning: Application of machine learning and deep learning methods for classification and detection of UAVs based on radar data [15, 17, 19, 20].
3. Micro-Doppler signature analysis: Using micro-Doppler analysis to identify and classify UAVs based on their motion and characteristics [12, 14, 16].

4. Statistical recognition: Using statistical methods to recognize and classify UAVs based on radar signals [6, 9, 16].
5. Use of radar spectrograms: Classification of UAVs based on radar spectrograms and analysis of their characteristics [15, 17, 20].
6. Use of multi-frequency radars: Classification and detection of UAVs using multi-frequency radars and analysis of their micro-Doppler signatures [12, 14].

The authors of articles [15, 18] use deep neural networks (DNN) to classify UAVs, but with different goals and approaches. The paper [15] focuses on UAV classification based on radar spectrograms using convolutional neural networks (CNN). The focus is on analyzing radar data to identify and classify UAVs.

The operating principle of the technology includes the following stages:
1. Signal recording: Signals from various targets, including unmanned aerial vehicles, are recorded using FMCW radar.
2. Conversion to spectrograms: Recorded signals are converted to spectrograms via Short-time fourier transfer (STFT), which allows visualization of micro-Doppler effects that occur when targets move.
3. Analysis using deep learning CNN: Spectrogram characteristics are analyzed using convolutional neural networks such as ResNet to classify and identify targets based on micro-Doppler signatures.
4. Development of ResNet-SP model: A ResNet-SP model is created with reduced computation but increased accuracy and stability, which facilitates efficient target classification.
5. Model training and performance evaluation: Models are trained on a proprietary radar spectrogram dataset and then their performance is evaluated in real time.

Thus, the authors' technology [15] is based on the use of radar spectrograms and deep learning to effectively classify and identify UAVs based on their micro-Doppler characteristics. The article [18] proposes a hybrid DNN model to classify the intent of unmanned aerial vehicles based on radar data. Here the focus is on determining the intentions and targets of the UAV based on telemetry and radar measurements. The operation of the proposed system can be described in three stages:

- data preparation and radar simulation;
- creation of a hybrid deep neural network;
- network training and testing.

A technology based on RCS analysis for statistical recognition of UAVs at microwave frequencies was proposed by the authors [6]. The main stages of operation of the proposed radar system:
1. UAV RCS measurement: Carrying out RCS measurements of various commercial UAVs at specific frequencies in a special compact band anechoic chamber to obtain information about the reflectivity of target UAVs;
2. RCS Analysis: Analyze RCS data and fit it to various statistical models using maximum likelihood, AIC [define acronym] information criterion and Bayesian information criterion (BIC) methods to select the best statistical models for each type of UAV;
3. Statistical UAV Recognition: Development of a UAV recognition system that estimates the probability of RCS test data belonging to one of the UAV types in the database using the most appropriate statistical distribution selected by AIC and BIC criteria to classify unknown UAVs based on their RCS;
4. Performance Evaluation: Conduct Monte-Carlo analysis to evaluate the effectiveness of the technology and study the influence of signal-to-noise ratio, frequency and polarization on the accuracy of UAV classification.

The system proposed by the authors [13] is based on the use of a bistatic radar operating at 5G frequencies. Bistatic radar uses two separate devices to transmit and receive signals, allowing it to effectively detect objects that reflect signals. The radar analyzes reflected 5G signals that do not pass through the line of sight and are reflected from the UAV body. These reflected signals, known as NLOS (non-line-of-sight), help determine the presence of a drone in the surrounding area. Additionally, the system captures received signal strength indicator (RSSI) values, which provide information about the drone's signal strength at various locations. This allows you to determine where the drone is and what its signal strength is at a specific point. The K-Nearest Neighbors (KNN) machine learning algorithm is used to predict the presence of a drone and its location in a 2x2 m grid. In order to evaluate the performance of the system, experiments were carried out both indoors and outdoors. The results showed that this technology can effectively detect and localize drones. The performance of radar systems for detecting and classifying UAVs based on the technologies described above is presented in Table 1.

Table 1
Performance of radar technologies

| Link | Technology | Performance (Precision) |
|---|---|---|
| [6] | RCS analysis for statistical recognition of UAVs | At SNR 10 dB: 97.73% |
| [7] | Using bispectral slices and the GA-BP neural network | At SNR 3 dB: 80% |
| [10] | NeXtRAD multistatic radio system | At SNR 6 dB: 92.5% |
| [11] | Machine learning and analysis of FMCW radar data | - |
| [12] | Radars of various frequency ranges, machine learning | 85.1% |
| [13] | Bistatic radar operating at 5G frequencies | - |
| [15] | Deep learning using radar spectrograms that reflect micro-Doppler effects | Availability of UAV: 100% |
| [16] | Combination of radar data and CNN | UAV localization: |
| [17] | Analysis of target movement characteristics and the use of machine learning methods | 75% |
| [18] | Hybrid deep neural networks and non-cooperative radars | 98.97% |
| [19] | Radar platform with FMCW MIMO technology | At SNR 10 dB: 80%; |

### *2.2. UAV detection and recognition technologies based on RF sensors*

RF-based systems for UAVs detection and classification (Fig. 3) are capable of operating with high accuracy by integrating the following technologies:

- Passive RF sensing [22];
- Compressive sensing [27];
- Machine learning and Deep learning [25];
- RF fingerprints [21, 26, 28, 29];
- Feature generation (FEG) [24].

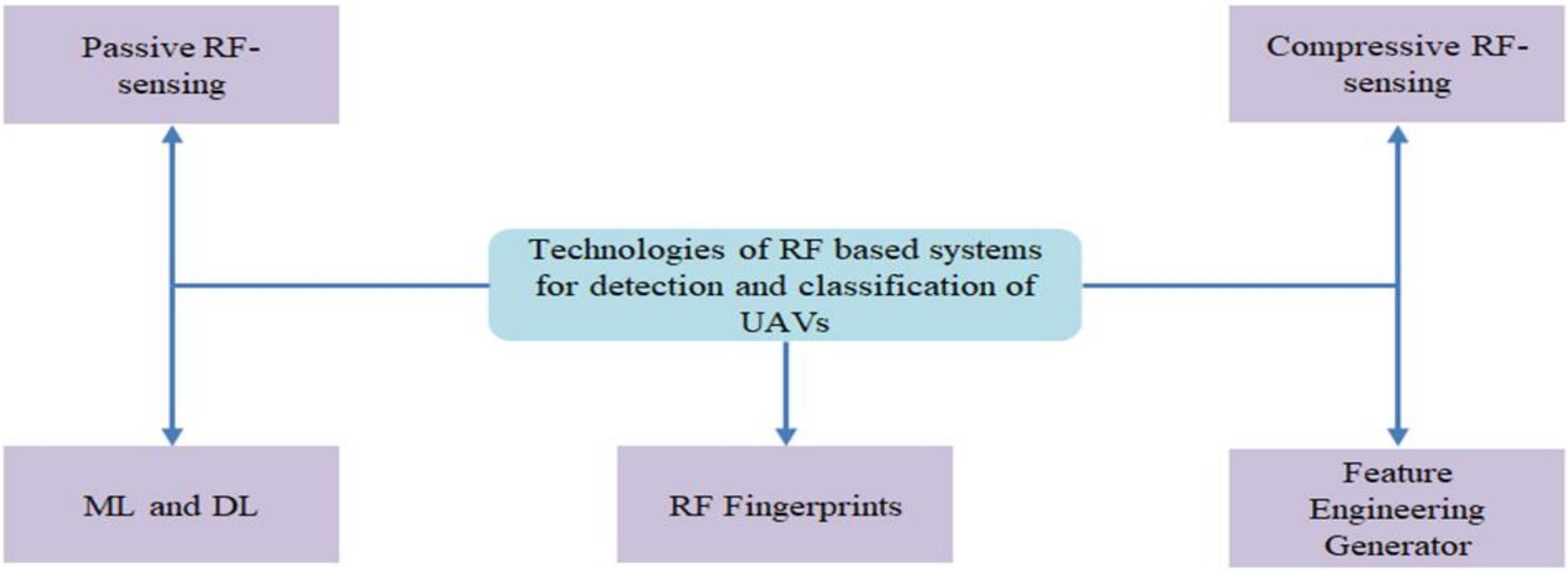

Fig. 3. Diagram of RF-based systems for detecting and classifying UAVs.

Passive RF sensing relies on analyzing RF signals and using machine learning to recognize preamble patterns, reducing computational costs. Thus, the system is capable of acting as a stand-alone intrusion detector or as part of a distributed sensor network. The drone presence detection process is carried out on-board the embedded computer, and the identification of the drone model can be performed in the data fusion center. Compressive sensing is used to process radio frequency signals to efficiently extract information from these signals with a minimum number of measurements. Deep learning is used to build a neural network capable of detecting and classifying UAVs based on these compressed RF signals. The authors' technology [22] is based on the use of passive RF sensing using a distributed network of sensors to detect and identify UAVs. The system includes several sensors that collect RF signals emitted by drones. The signals are sent to a data center where software-defined radio (SDR) is used to analyze and process the RF signals. Hardware-accelerated time-frequency computing allows for efficient real-time signal processing and identification of UAV-specific characteristics. UAV detection and identification algorithms are configured for high accuracy and low false alarms. After processing the data, the system is able to detect the presence of a UAV, classify its model and transmit information to the operator to take appropriate action. In the article [25], the authors proposed a technology with the development of machine learning based on RF signals. By extracting spectral characteristics from RF signals, a unique RF signature is created for each type of UAV. The XGBoost machine learning algorithm is then applied to analyze these RF signatures to determine the presence of a UAV, its type and mode of operation.

The technology process includes the following steps:

- Extracting spectral characteristics from radio frequency signals that are emitted by UAVs;
- Creation of a unique RF signature for each type of UAV based on the extracted spectral characteristics;
- Training machine learning models using the XGBoost algorithm based on unique RF signatures;
- Evaluation of models for the accuracy of detecting the presence of a UAV, determining its type and operating mode.

A new approach combining Feature Engineering Generator (FEG) and Multi-Channel Deep Neural Network (MC-DNN) methods for detecting and classifying UAVs based on RF signals was proposed by the authors of [24]. This technology includes four stages. First stage: Collection and pre-processing of RF signals. At this stage RF signals received from the UAV are collected. Signals undergo pre-processing, including noise filtering, data normalization and other methods to improve their quality, and prepare for further analysis. Second stage: Feature Engineering Generator (FEG). This step involves extracting key features from preprocessed RF signals, including separating different frequency components, reducing signal similarity, and extracting unique characteristics for each signal type. FEG helps make signals more distinguishable and informative for further analysis.

Third stage: Using Multi-Channel Deep Neural Network (MC-DNN). At this stage, multi-threaded (multi-channel) classification of RF signals occurs based on the extracted features that were obtained using FEG. MC-DNN is trained to recognize different types and flight modes of UAVs. Fourth stage: Detection and classification of UAVs. At this stage, after processing and classifying RF signals using MC-DNN, the presence of unmanned aerial vehicles is detected and classified into types and flight modes. The performance of radio frequency UAV detection and classification systems based on the functionality of the above technologies is presented in Table 2.

Table 2

Performance of RF-based UAV detection and classification systems

| Link | Technology | Performance (Precision) |
|---|---|---|
| [22] | Passive RF sensing | 99.94% at range up to 150m;<br>99.38% at range up to 350m |
| [24] | FEG and MC DNN | Precision: 98.4%,<br>F1: 98.3% |
| [25] | Machine learning (XGBoost) | UAV detection: 99.96%<br>Type of UAV detection: 90.73% |
| [26] | Sequential CNN | Precision: 92.5%<br>F1: 93.5% |
| [27] | DNN, CNN | 99% |
| [28] | DNN | At SNR 0 dB: 99% |
| [29] | RF fingerprints | 99% |

### *2.3. Optical-electronic systems for UAVs detection and classification*

Optical-electronic systems for UAVs detection and recognition can be classified according to technical means and classification technologies.

Classification by technical means:

- systems based on optical surveillance cameras [30, 31, 33, 34, 35, 36];
- optical-electronic systems based on thermal imaging infrared sensors [38];
- UAV detection and classification systems based on laser-optical sensors (LiDAR) [39].

The following object recognition technologies are also subject to classification:

- Deep learning [30, 31, 32, 33, 34, 35, 36, 37, 40];

Machine learning [30, 34].

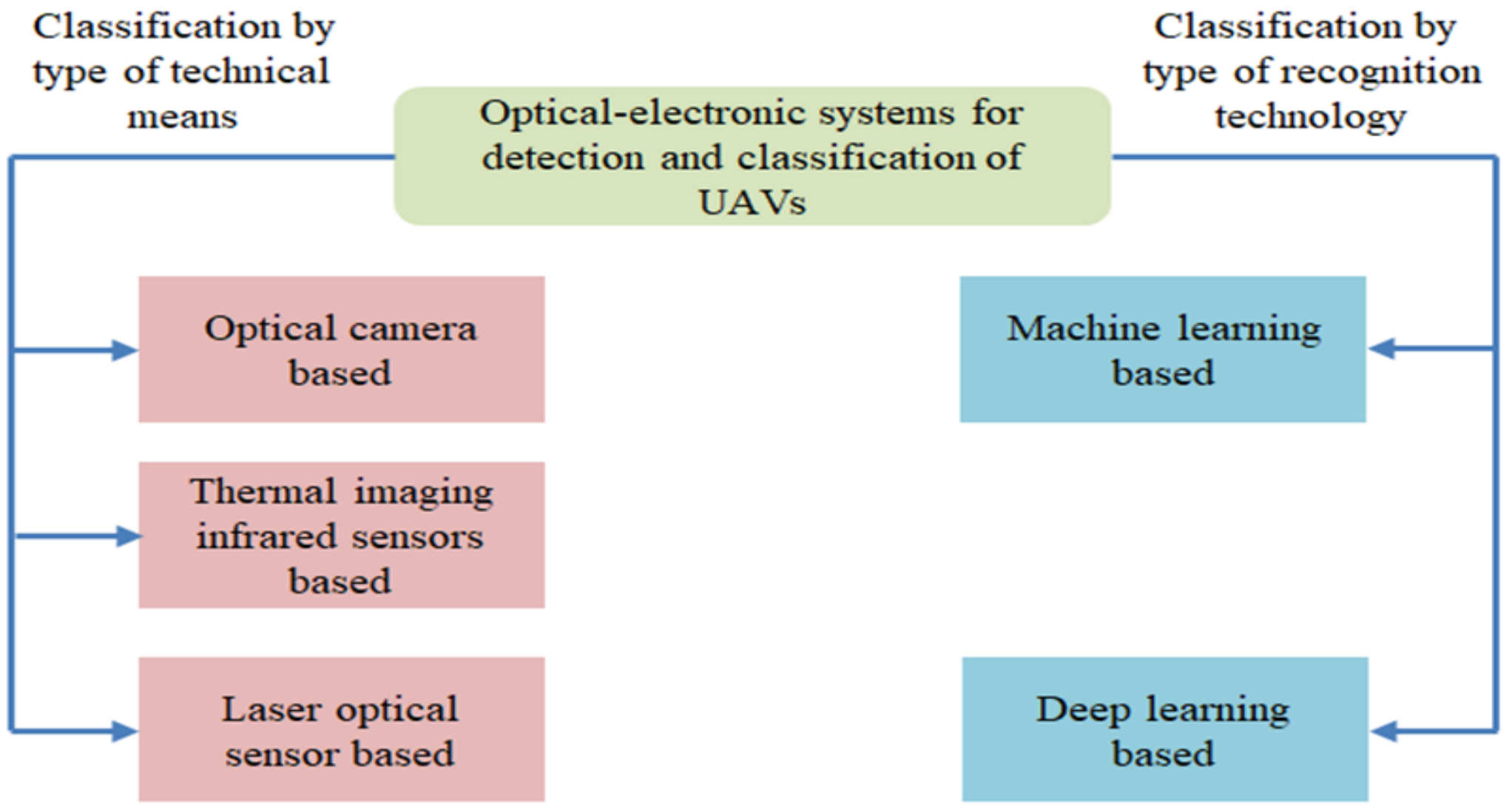

Fig. 4. Diagram of optical-electronic systems for detecting and classifying UAVs.

The authors of [30] present a method for drone detection using CNN in comparison with support vector machines (SVM) and k-nearest neighbors (KNN). This work proposes two stages.

1. Drone image feature extraction: This step involves analyzing UAV images to extract key features that may be important for their classification. These cues may include the UAV's shape, size, texture, color, and other characteristics that help differentiate the drone from other objects.
2. Training a convolutional neural network on the extracted features: After extracting the key features of the drone images, this data is used for training. The CNN is trained on reference data to learn to recognize and classify drone images based on the extracted features.

The testing phase includes two stages:

1. Classification of new drone images. The trained CNN model is applied to classify the new input UAV images. The model uses the learned features to determine whether the image is a UAV or not.
2. Improving UAV detection precision using CNN: CNN helps distinguish UAV images from bird images, which helps improve UAV detection accuracy. By learning from extracted features and the ability to identify complex patterns, CNN can effectively distinguish objects and improve classification precision.

The results of the research [30] show that CNN outperform SVM and KNN in UAV detection, achieving an accuracy of 95%.

The article [34] presents a technology for detecting UAVs in video with high accuracy in real time. The authors divide the UAV detection task into two stages: detecting moving objects and classifying them into drones, birds or background.

1. Detection of moving objects. To identify moving objects, the background proofreading method is used, which allows you to select objects that differ from a static background. This method helps to quickly and accurately identify objects that move in a video stream.
2. Classification of objects. Detected objects are classified using a convolutional neural network (CNN) that is trained on a set of images of UAVs, birds, and backgrounds.

Based on the YOLOv5 deep learning model, an autonomous UAV detection system was implemented in [35]. The operating principle of this system includes 4 main components.

1. YOLOv5 algorithm: object detection algorithm that works in real time. It divides the image into a grid and, for each grid cell, predicts bounding boxes and object class probabilities. One of YOLO's key advantages is its ability to detect objects quickly and accurately.
2. Model training: To train the model, the authors used pre-trained weights and data augmentation techniques. This allows the model to train on a small data set and prevents overfitting. The model is trained on labeled images of UAVs to learn how to recognize them in the images.
3. Autonomous detection: Once trained, the model can operate autonomously, detecting UAVs in images without the need for human intervention. This allows the technology to be used to automate the UAV detection process in real time.
4. Evaluation of results: After training, the model is evaluated using metrics such as average precision (mAP) and recall to measure its performance and accuracy in UAVs in images.

The results show that YOLOv5 outperformed previous versions of YOLO in UAV detection speed and precision. The technology presented in [38] is based on the integration of three key elements to improve the quality of UAV detection and classification through thermal image analysis.

1. Saliency Cues. Calculating a highlight map helps highlight areas where UAVs are likely to be located in thermal images. This extraction map is combined with the feature map of the Faster-RCNN neural network to improve feature extraction and model training.
2. Sign of enlarged small objects (ESO). The ESO module is designed to magnify small drones in the input image, which helps the neural network detect them better.
3. Data Augmentation Technique. The proposed data augmentation technique makes it possible to create new thermal imaging images of drones from various sources for training a neural network.

The above components work together to provide more accurate and reliable UAV detection in a variety of scenarios and lighting conditions. The technology described in [40] is based on the use of a modified version of the YOLOv8 network for detecting small UAVs.

The authors present an improved detection model that includes additional modules and mechanisms to improve the precision and efficiency of detection. The model architecture introduces an attention mechanism to improve target feature fusion and overall performance for UAV detection. The performance of optical-electronic UAV detection systems described in [30 – 40] is presented in Table 3.

Table 3

Performance of optical-electronic systems for detecting and classifying UAVs

| Link | Detection and classification technology | Performance (Precision) |
|---|---|---|
| [30] | CNN | 95% |
| [31] | YOLOv4 | mAP 0.7436<br>F1-measure 0.79 |
| [32] | YOLOv2 | mAP 0.7497 |
| [33] | YOLOv2 | - |
| [34] | Background proofreading and CNN | 70.1%, F1- measure 0.742 |
| [35] | YOLOv5 | mAP 90.40% |
| [36] | YOLOv8 | - |
| [37] | YOLOv4 | IoU - 84%<br>mAP - 83%<br>Precision -83% |
| [38] | saliency cues and magnified features on thermal images | AP – 0.9498<br>DD – 0.852 |
| [39] | LiDAR point clouds | increasing range to 100 m. |
| [40] | YOLOv8<br>YOLOv7-tiny<br>YOLOX-s<br>YOLOv5-s | mAP – 95.3%<br>mAP – 85%<br>mAP – 88.7%<br>mAP – 91.2% |

## *2.4. UAV detection and classification technologies based on acoustic sensors*

Acoustic systems for detecting and classifying UAVs are divided according to the principle of operation into passive, active and combined (fig. 5).

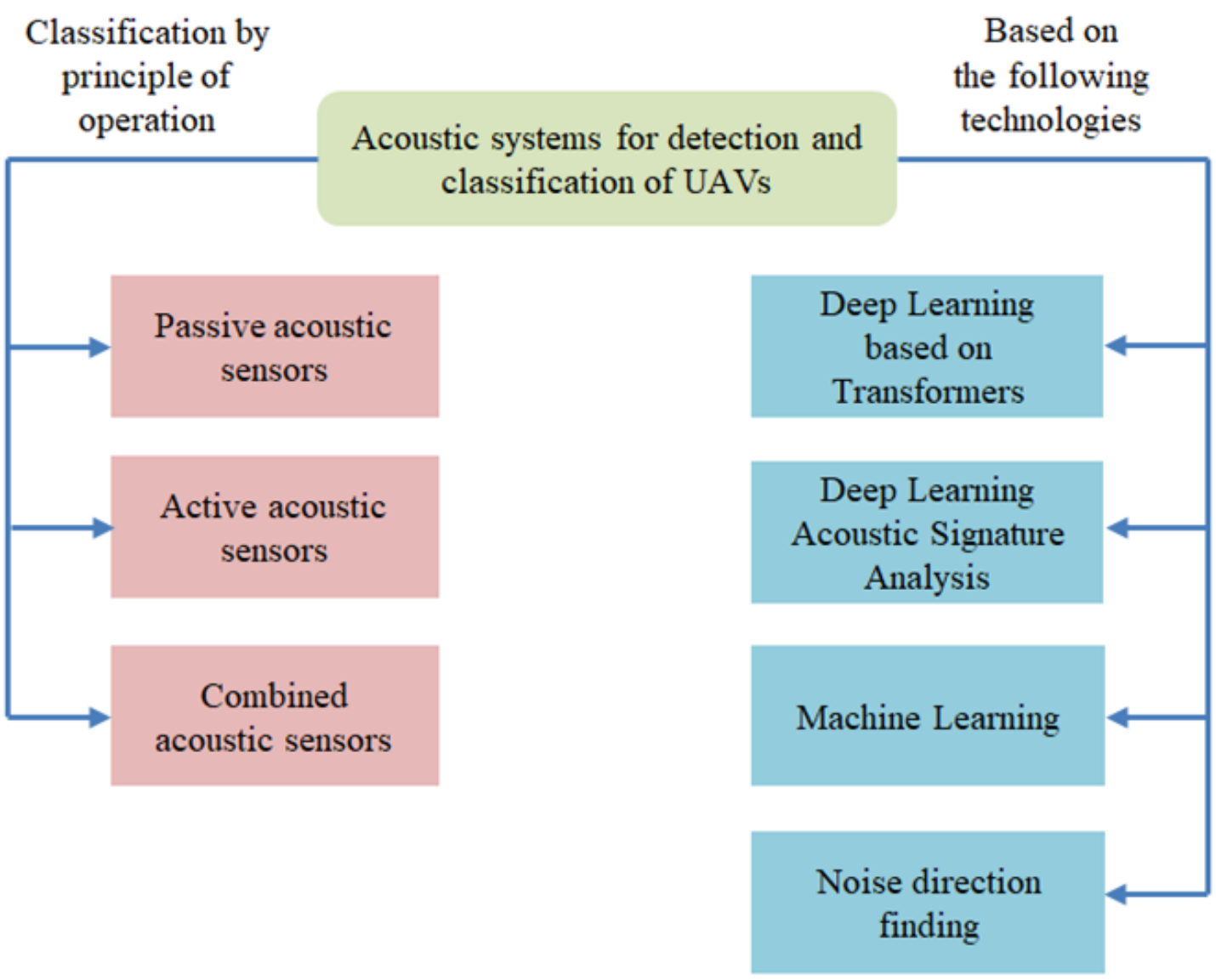

Fig. 5. Diagram of acoustic systems for detecting and classifying UAVs.

Passive acoustic systems use microphones to detect the sound produced by the UAV's engine. These systems can be useful for detecting UAVs over long distances, but they cannot identify the type or model of the UAV. Active acoustic systems use transmitters to emit sound waves that bounce off the UAV and return back. These systems can detect the type and model of UAV, but they have a limited range. Combination systems combine passive and active acoustic systems to provide both detection and recognition of UAVs. These systems typically have a long range and can identify the type and model of UAV. It is important to note that the effectiveness of these systems can depend on many factors, including environmental noise, weather conditions and the characteristics of the UAV itself. Analyzing algorithms for detecting and classifying UAVs as targets based on acoustic sensors, we can identify a number of technologies that provide high performance in accordance with the criteria of precision and range:

- Deep Learning using transformer-based models [49];
- Deep Learning based on the analysis of acoustic signatures [42, 43, 44, 46, 47];
- Machine Learning [45, 48];
- Noise direction finding (sound intensity analysis, spectral analysis and direction finding) [41].

The research conducted by the authors in [44] focused on developing UAV detection and identification system using acoustic signatures and deep learning techniques. This paper presents an innovative solution for automatic detection and identification of UAVs based on their acoustic characteristics.

The operating principle of the system can be divided into four stages:

1. UAV detection. The system first detects the presence of a UAV in a certain area using acoustic sensors.
2. UAV identification. The system analyzes the characteristics of the UAV, such as its type or model, to determine whether the object is authorized or unauthorized.
3. UAV localization and tracking. The system determines the location of the drone and tracks its movement.
4. Obstructing the UAV mission. In the final stage, the system takes measures to interfere with the UAV mission using various methods such as shooting at the drone, using networks, using a laser beam, or interfering with the communication between the controller and the UAV to land it safely.

The article [49] presents an innovative approach that allows not only to detect a UAV, but also to recognize possible anomalies in its operation. The principle of operation is to use acoustic data collected using a microphone mounted on the UAV. This data is then put through an analysis process using DNN, including transformer and CNN models. Table 4 shows the performance indicators of modern acoustic systems for detecting and classifying UAVs.

Table 4

Performance of acoustic UAV detection and classification systems

| **Link** | **Technology** | **Performance (Precision)** |
|---|---|---|
| [43] | LD | 63.25% |
| | MLP | 59.32% |
| | SVM | 65.52% |
| | RF | 51.01% |
| | YAMNet | 71.23% |
| [44] | Machine Learning | 99% |
| [45] | Machine Learning | Recall: 99.8%<br>Precision: 100% |
| [46] | CNN | 96.3% at a range of 150m for small class UAVs, and 500m for medium and large UAVs |
| [47] | CNN | 92.47% |
| [48] | Deep learning with transformer and CNN models | 88.4% |

### *2.5. UAV detection and classification technologies based on Sensor Fusion*

Detection and classification of UAVs using hybrid technologies Sensor fusion is the integration of the functionality of different types of sensors in order to improve system performance in terms of precision, range and probability of false alarms. Based on data analysis, Sensor Fusion systems are able to identify UAVs in relation to objects of another class (in particular birds), determine the type, its unique characteristics, current dynamic parameters (speed, direction, range, etc.). This technology is used for various purposes, including detection and tracking of UAVs, preventing unauthorized flights, ensuring airspace security and others.

As a result of a review of scientific articles from the Scopus and Web of Science databases not older than 2020, the following combinations of Sensor Fusion technologies were identified:

- acoustic and optical-electronic sensors [51, 55, 56, 57];
- optical-electronic satellite sensors and radars [50, 51, 54, 59, 60];
- optical-electronic, acoustic sensors and radars [52], [62];
- optical-electronic and RF-based sensors [55];
- optical-electronic, RF-based and acoustic sensors [53].
- RF-based and acoustic sensors [61].

In [56], the authors proposed a multi-sensor UAV detection system using advanced machine learning methods. The operating principle of this system is to combine data from various sensors for detection and classification. The system operation algorithm is presented in the form of five stages:

1. Data collection. Various sensors such as an infrared camera, a video camera, audio sensors and a panoramic camera collect information about the surrounding environment.
2. Data processing. The obtained data is processed using machine learning methods to analyze and extract features characteristic of the UAV.
3. Data fusion. Data from various sensors is combined to create a comprehensive overview of the surrounding environment and detect potential UAVs.
4. Classification. Based on the extracted features and data from various sensors, the detected objects are classified as UAVs or other objects (for example, birds, airplanes).
5. Tracking. The system can also track the movement and trajectory of detected objects for further analysis and monitoring.

The results of the study [56] indicate the successful application of general machine learning methods to infrared sensor data, making them effective for UAV detection. The infrared detector achieves an F1-measure of 0.76 (76%), which also confirms the good Precision and Recall of UAV detection and classification. This result is comparable to the performance of the video detector, which has an F1-measure of 0.78 (78%) and the audio detector of 0.93 (93%). Thus, the use of infrared sensors, audio sensors, video detectors (video surveillance cameras) in combination with machine learning methods can be considered an effective way to detect and classify UAVs, providing high performance according to accuracy criteria (F1-measure, Recall).

In article [55] the authors proposed a detection system based on artificial neural networks, where a DNN processes data from RF-based sensors, and a CNN processes images from a video detector. Features from both networks are combined and fed into the input of another DNN, which produces an assessment of the presence of a UAV. The accuracy of this technology was 75%.

A data fusion strategy based on combining frames of optical-electronic surveillance equipment and radar signals to improve UAV detection was proposed by the authors [50]. This Sensor Fusion system has two levels:

1. Data layer. At this level, the information received from the optical-electronic sensor (video detector) and radar is merged to determine the coordinates and position of the UAV. Optical-electronic data is used to extract visual characteristics of a target, while radar signals can help determine the distance and speed of an object.
2. Decision-making level. At this level, the unique characteristics of the UAV are merged to more accurately identify and classify the target. Information about the UAV visual characteristics and coordinates is combined to make a target identification decision.

Thus, the technology [50] consists of the joint use of data from computer vision and radar to determine the coordinates and unique characteristics of the UAV.

In [61], the authors presented a technology for fusion of RF and acoustic characteristics for UAV classification. They propose using both classical and deep machine learning methods, as well as combining RF and acoustic features to more accurately and efficiently classify drones. Their study used various neural network-based classifiers (DNN, CNN, LSTM, GRU) and the classical SVM method. The accuracy of this technology is about 91% at a signal-to-noise ratio (SNR) of -10 dB using LSTM features. Its advantage is high noise immunity and significant improvement in accuracy in noisy environments with low signal-to-noise ratio. In addition, this technology outperforms existing drone classification methods, especially in low signal-to-noise ratio (-20 dB) environments.

In [62], the authors presented a sensor fusion strategy for non-cooperative detection and avoidance (SAA) for UAVs flying at low altitudes. This strategy integrates radar and optical-electronic sensor data at both the detection and tracking levels to benefit accuracy. The proposed strategy is called Fuse-Before-Track (FBT) through a two-step approach of using radar and visual information together to remove uninteresting echoes and ground clutter. Tracking is also accomplished by using radar measurements and combining visual detections to improve the accuracy of the solution. The relative positioning decision is made using differential carrier phase GNSS techniques. Thus, the improved layer fusion approach provides meter-level and meter-per-second level radar detection accuracy for range and speed, respectively.

### 2.*6. Use of UAV detection and classification technologies in modern Anti-Drone systems*

Currently, many companies are designing Anti-Drone devices and systems that are based on the optical-electronic, acoustic, RF and radar technologies for detecting and classifying UAVs discussed above. These software and hardware systems combine several technologies, corresponding to the well-known Sensor Fusion principle. The optical-electronic components of Anti-Drone systems use cameras and infrared sensors for visual detection. The radar unit is aimed at emitting electromagnetic waves, which are reflected from objects and return back to the radar, allowing you to determine the distance to the object, its speed and direction. The radio frequency detection channel of Anti-Drone systems monitors the radio frequency spectrum for the presence of signals transmitted by drones. Acoustic sensors in the systems detect drones by the sound they make. An indispensable component of the Anti-Drone systems software package is artificial intelligence. Artificial intelligence is represented by neural networks, machine learning and deep learning algorithms that perform classification tasks, UAV recognition, data analysis from different sensor channels, and decision making. In addition to detection and classification tasks, Anti-Drone systems are capable of eliminating UAVs.

Common methods of suppressing and destroying UAVs:

- Electronic jamming - this method includes cyber-attacks, GPS substitution, interception of control and jamming, which contributes to the loss of communication between the drone and the operator or to his disorientation;
- Physical impact - interception of a UAV by a fighter drone, the use of nets and projectiles;[63];
- Powerful microwave radiation - damage to UAV electronics by powerful microwave radiation, which leads to failure of electronic components [63];
- Laser – the use of a laser as a weapon [64] or as a countermeasure to the optical sensors of UAVs [65].

The work [66] presents destructive and non-destructive methods of UAV cyberattacks. Non-destructive cyber-attack methods include four approaches. The first approach is information leakage, which can lead to the disclosure of protected information. An example would be obtaining information about a UAV's mission or data to look up a Wi-Fi password. The second approach is integrity violation, where UAV data may be destroyed, corrupted, or altered. This type of attack can only be detected after it has occurred. The third approach is denial of service (DoS), which makes it impossible to use the UAV system. The fourth approach is illegal use of information aimed at accessing non-public parts of the system or using certain services without permission. For example, using a stolen password to decrypt an intercepted communication or seize control. These cyberattack techniques can result in obtaining sensitive information, altering data, or disrupting a system without causing physical damage. Destructive methods of cyber-attacks involve a direct impact on the object, which ultimately leads to irreversible physical damage to the UAV hardware complex, due to vulnerabilities at low levels of the OSI model.

Based on quality parameters, such as detection efficiency, identification accuracy, reliability, speed, information security, adaptability and flexibility, the following Anti-Drone systems can be distinguished:

- DedroneRapidResponse [76];
- IXI Technology Drone Killer [71];
- SkyWall Patrol [67];
- Liteye AUDS [74];
- Kaspersky Antidrone [77];
- MSI-Defence Systems Terrahawk Paladin [78];
- Elbit Systems ReDrone [79].

DedroneRapidResponse provides a mobile solution for detecting, tracking and identifying UAVs based on artificial intelligence. This software and hardware complex uses cloud software to accurately determine the location of the drone even when the target takes off, which gives security teams and special units a significant gain in time. DedroneRapidResponse does not include built-in tools for neutralizing UAVs. However, it can be integrated with other systems to provide a full range of protection. Featuring a portable drone detection subsystem, including radars, RF sensors and an optical-electronic recognition channel, DedroneRapidResponse is capable of detecting and identifying UAVs at a distance of more than 5 km.

The IXI Technology DroneKiller Anti-Drone system is a portable device that uses SDR technology to detect and eliminate UAVs. This device blocks GPS signals, forcing targets to descend or return to their take-off point. DroneKiller is also capable of disabling UAVs on seven frequency bands at a distance of up to 1000 meters. Due to its portability and lightness, this Anti-Drone device can be deployed inside light vehicles, as well as used by first response teams in mobile units, strike groups, and checkpoints.

The Liteye Anti-UAV Defense System (AUDS) is a strategic counter-UAV system designed to detect, track and neutralize them. AUDS integrates radar scanning for target acquisition, electro-optical tracking and classification, and directional radio frequency jamming technology. The system is used to protect critical infrastructure in dense urban areas. The installation is also deployed on mobile and robotic vehicles to support a range of missions such as air and ground base defense, convoy protection, offensive electronic attack and non-military missions such as wildfire response and event security.

The SkyWall Patrol Anti-Drone is a portable system that provides the operator with the ability to physically capture the drone in a specially designed network. The device uses a built-in SmartScope optical-electronic sight for detection and targeting of UAVs. The sight automatically compensates for the UAV's speed and distance, allowing the operator to accurately target it. After identifying a target, SkyWall Patrol uses compressed air to launch a projectile towards the UAV. The projectile splits into two as it approaches the target and deploys a net to capture it. Some projectiles contain a parachute that controls the descent of the captured UAV, minimizing the risk of collateral damage and keeping the device intact for later analysis.

SkyWall Patrol is not only used as a standalone anti-UAV solution, it also integrates with the wider security system using the SkyLink module to create a highly effective countermeasures package. A single SkyWall Patrol system can protect a specific area, or multiple systems can be deployed from mobile platforms to protect a large strategic location.

Kaspersky Anti-Drone is a software and hardware system that uses artificial intelligence and neural networks to detect, classify and track UAVs of various types and classes. This system uses a primary detection module, which may be a device capable of scanning the airspace above the protected area. Such a device can be a lidar, radio frequency sensor, radar, or PTZ video surveillance camera. Artificial intelligence-based algorithms are used to recognize and classify UAVs. This allows for high speed data processing from hardware sensors of various types and provides comprehensive analytics for UAVs in a single interface. Kaspersky Anti-Drone also uses an optical detection module that classifies the drone at any time of the day thanks to PTZ cameras. After detecting and classifying the UAV, the system tracks the target, waiting for a command to be sent to the jammer.

Table 5 provides information on known models of Anti-Drone systems, the main characteristics and technologies used for detecting, classifying and eliminating UAVs.

Table 5

Performance of acoustic UAV detection and classification systems

| System description | Characteristics (range, ranges, accuracy) | Detection and classification technologies | UAV elimination technologies |
|---|---|---|---|
| SkyWall Patrol [67] | Maximum Range: 100 meters (330 feet) | Optical-electronic | Physical elimination |
| DroneShield DroneGun Tactical [68] | Maximum Range: 2000 meters | RF-based | Electronic jamming |
| Aaronia Drone Detection System [69] | Maximum Detection Range: Up to 50 kilometers. Frequency Range: 20 MHz to 8 GHz. Tracking Accuracy (line of sight): 4 to 6 degrees. | RF-based | Electronic jamming |
| Fortem DroneHunter [70] | Detection Range: 500 meters. Azimuth and Elevation Accuracy: ±2° | Radar | Physical elimination |
| IXI Technology Drone Killer [71] | Effective Range: Up to 1000 meters or approximately 0.62 miles | RF-based | Electronic jamming |
| DJI Aero Scope [72] | Reliable Detection Range: Up to 16 km. Maximum Detection Range: Up to 32-48 km. Maximum Monitoring Range: Up to 50 km. | RF-based | - |
| Anduril Lattice Counter Drone System [73] | Maximum detection range for UAV: from 5 to 20 km (depends on UAV size) | Radar, optical-electronic | Physical elimination (shot down by another UAV) |
| Liteye AUDS [74] | Detection Range: Up to 10 km. Effective Range for Micro UAVs: Up to 2 km. | Radar, optical-electronic | Electronic jamming |
| ELI-4030 Drone Guard [75] | Detection Range: Up to 4.5 km. DOA Accuracy: ±2°. Frequency Coverage: 400 to 6000 MHz. | Radar, RF-based | Electronic jamming |
| DedroneRapidResponse [76] | RF Range: Up to 5 km. PTZ Range: Up to 1 km. Radar Range: Up to 1 km. | Radar, optical-electronic, RF-based | - |
| Kaspersky Anti-drone [77] | Detection Range: Up to 5 km. RF Range: Up to 1 km. Average Classification Accuracy: 97%. | Radar, optical-electronic, RF-based | - |
| MSI-Defence Systems | Elimination range of air targets: | Radar, optical-electronic | Physical elimination (use of |
| Terrahawk Paladin [78] | up to 2 km. | | automatic artillery mount) |
| Elbit Systems ReDrone [79] | Range: from 2 km or more; 360° coverage | Radar, optical-electronic, acoustic, RF-based | Electronic jamming |

The MSI-Defence Systems Terrahawk Paladin [78] system is a counter-UAV system that provides target elimination in the short range of air defense. This Anti-Drone consists of three main components: a radar station, a digital fire control system and an automatic artillery mount. The system uses FIELD and SKY radars, as well as AI technologies for ballistic calculations. An optical-electronic unit with daytime and thermal imaging cameras helps identify potential threats during the day, in adverse conditions or at night. The Mk 44 Bushmaster II 30mm artillery mount is used to destroy targets. When using ammunition with software detonation, the range of destruction of air targets can reach up to 2 km.

The Elbit Systems ReDrone [79] system is an innovative solution for detecting and blocking threats associated with the use of UAVs. It is designed to ensure the security of military, police, civilian facilities, as well as at state borders, airports, seaports, strategic sites, public events, landmarks, prisons, military bases and convoys. ReDrone uses a unique set of countermeasures that detect, identify, classify, locate, track, neutralize, manipulate and destroy multiple threats simultaneously. Various detection technologies are used for this, including SIGINT, which detects all types of commercial drones, identifies them and alerts the operator. 2D or 3D radars are also used for active detection and positioning. Once a target is detected, ReDrone blocks all radio and GNSS signals, disrupts the drone's communication with the operator and sends it off course, preventing the possibility of an attack.

Thus, the future of modern Anti-Drone systems is determined in the advancement of defense technologies, the application of artificial intelligence and neural networks, and integration with existing security systems. These systems are being developed to detect, classify, and mitigate UAVs using a variety of methods, including camera systems, drone detection radar, net guns, and cyber takeover systems. The use of Sensor data fusion, such as RF-based analyzers, acoustic sensors, optical sensors, and radar, is also a significant area of focus. As the technology evolves, these systems are expected to become more efficient and accurate, providing a robust solution to the growing threat posed by UAVs.

## 3. Conclusion

In conclusion, the field of UAV detection and classification is advancing rapidly, driven by the integration of diverse technologies such as radar, RF sensors, acoustic detectors, and optical imaging systems. The incorporation of machine learning and artificial intelligence plays a pivotal role in refining these technologies, enabling more accurate and reliable systems. Sensor fusion technology, which combines data from multiple sources, exemplifies the trend towards holistic detection approaches that can adapt to the varied challenges presented by UAV monitoring. Continued innovation is essential to address the evolving complexities of UAV technologies and their applications. By improving the sensitivity and specificity of detection methods and harnessing the power of AI for real-time analysis and decision-making, future systems will not only enhance airspace security but also facilitate the safe integration of UAVs into civil and commercial environments. This comprehensive progress in UAV detection and classification technologies holds significant promise for a wide range of applications, making ongoing research and development a crucial investment for both safety and technological advancement.

## 4. Acknowledgments

This research is funded by the Science Committee of the Ministry of Science and Higher Education of the Republic of Kazakhstan (Grant No. AP19679009).

**Declaration of interests**

☐The authors declare that they have no known competing financial interests or personal relationships that could have appeared to influence the work reported in this paper.

☒The authors declare the following financial interests/personal relationships which may be considered as potential competing interests:

Ildar Kurmashev reports financial support was provided by Manash Kozybayev North Kazakhstan University. This research is funded by the Science Committee of the Ministry of Science and Higher Education of the Republic of Kazakhstan (Grant No. AP19679009) If there are other authors, they declare that they have no known competing financial interests or personal relationships that could have appeared to influence the work reported in this paper.